\documentclass{article}

\PassOptionsToPackage{numbers, compress}{natbib}

\usepackage[preprint]{neurips_2025}
 \usepackage{amsmath}
\usepackage{fontawesome5}
\usepackage{wrapfig}
\usepackage[table,xcdraw]{xcolor}

\usepackage{amsmath}
\usepackage{amssymb}
\usepackage{mathtools}
\usepackage{amsthm}
\usepackage{microtype}
\usepackage{graphicx}
\usepackage{subcaption}
\usepackage{booktabs} 
\usepackage{multirow}
\usepackage{enumitem}
\usepackage{algorithm}

\usepackage{algorithmic}
\usepackage{multirow}
\usepackage[utf8]{inputenc} 
\usepackage[T1]{fontenc}    
\usepackage{hyperref}       
\usepackage{url}            
\usepackage{booktabs}       
\usepackage{amsfonts}       
\usepackage{nicefrac}       
\usepackage{microtype}      
\usepackage{xcolor}         
\usepackage[most]{tcolorbox}
\usepackage{fontawesome5}
\usepackage{hyperref}
\usepackage{array} 
\usepackage{listings}
\usepackage{graphicx}  
\usepackage{subcaption} 
\usepackage{float}      
\usepackage{fontawesome5}
\definecolor{lightblue}{RGB}{235,245,255} 
\definecolor{liautoblue}{RGB}{71,111,182} 
\definecolor{textred}{RGB}{128,0,0}
\usepackage{titlesec}
\titleformat{\section}
  {\large\bfseries\color{liautoblue}}{\thesection}{1em}{}
\titleformat{\subsection}
  {\bfseries\color{liautoblue}}{\thesubsection}{1em}{}
  \titleformat{\subsubsection}
  {\bfseries\color{liautoblue}}{\thesubsubsection}{1em}{}
\usepackage[most]{tcolorbox}
\newtcolorbox{liautoabstract}{
    colback=lightblue,
    colframe=white,
    boxrule=0pt,
    arc=2mm,
    left=4mm,
    right=4mm,
    top=5mm,
    bottom=5mm,
    enhanced, 
    before upper={\setlength{\parindent}{0pt}} 
}
\usepackage[most]{tcolorbox} 

\newtcolorbox{stepTitle}[1]{
    enhanced,
    colback=gray!5,    
    colframe=black!50,  
    boxrule=-1pt,
    arc=0mm,            
    left=2mm, right=2mm, top=1mm, bottom=1mm,
    fontupper=\small,  
    title=#1
}

\tcbset{
    stepbox/.style={
        enhanced,
        colback=gray!20,         
        colframe=gray!50,        
        boxrule=0.5pt,
        title={#1},              
        fonttitle=\bfseries,     
        left=2mm, right=2mm, top=1mm, bottom=1mm
    }
}

\newtcolorbox[auto counter]{case}[2][]{ 
    enhanced,
    colback=gray!5,
    colframe=black!70,
    coltitle=white,
    fonttitle=\bfseries\sffamily,
    fontupper=\small,
    arc=1.5mm,
    boxrule=0.5pt,
    title=Case study \thetcbcounter: #2, 
    left=1mm, right=1mm, top=2mm, bottom=2mm,
    label type=case, 
    #1               
}

\newtcolorbox{toolbox}[1]{
    enhanced,                 
    colback=gray!5,           
    colframe=black!70,        
    coltitle=white,           
    fonttitle=\bfseries\sffamily,
    fontupper=\small,
    arc=1.5mm,                
    boxrule=0.5pt,            
    title=#1,                 
    left=1mm, right=1mm, top=2mm, bottom=2mm
}

\usepackage{fancyhdr}
\usepackage{graphicx} 
\hypersetup{
    colorlinks=true,
    urlcolor=liautoblue,
}

\title{Evolving from Tool User to Creator via Training-Free Experience Reuse in Multimodal Reasoning}

\author{%
   Xintian Shen \thanks{Co-first author} \And Jiawei Chen\footnotemark[1] \And Lihao Zheng\footnotemark[1] \AND Hao Ma \thanks{Technique Leader} \\
  \And Tao Wei  \footnotemark[2]  \And Kun Zhan \thanks{Supervisor} \\
}

\begin{document}

\maketitle

\begin{liautoabstract} 

Existing Tool-Integrated Reasoning (TIR) models have effectively extended the question-answering capabilities of LLMs by incorporating external tools. However, real-world scenarios present numerous open-ended problems where fixed tools often fail to meet task requirements. Furthermore, the lack of self-optimization mechanisms means that erroneous tool outputs can mislead the LLM's responses. Additionally, the construction of existing tools entails significant manual effort, which consequently constrains their applicability. 
Recognizing that the reasoning traces of LLMs encapsulate implicit problem-solving capabilities, we propose UCT, a novel training-free framework that transforms agents from tool users to tool creators. This approach harvests reasoning experiences and distills them into reusable assets.
This method transforms the agent from a mere tool user into a tool creator, enabling adaptive tool creation and self-updating during the inference process. We also introduce a memory consolidation mechanism to maintain the tool library, ensuring high reusability of retained experiential memory for subsequent reasoning tasks. This novel automated tool construction paradigm continuously improves tool quality during reasoning, allowing the overall agent system to progress without additional training. Extensive experiments demonstrate that our method serves as a novel paradigm for enhancing the capabilities of TIR models. In particular, the significant performance gains achieved +20.86\%$\uparrow$ and +23.04\%$\uparrow$ on benchmarks across multi-domain mathematical and scientific reasoning tasks validate the self-evolving capability of the agent.

\vspace{3mm}
    {\color{liautoblue!30}\rule{\linewidth}{0.5pt}} 
    \vspace{2mm}

    \small 
    \renewcommand{\arraystretch}{1.3} 
\begin{tabular}{@{} l l @{}}
        {\faCalendar*} & \textbf{Last Update Date:} February 2, 2026 \\
        {\color{liautoblue}\faEnvelope} & \textbf{Correspondence:} {zhankun@lixiang.com} \\
    \end{tabular}\end{liautoabstract}

\begin{figure*}
    \centering
    \includegraphics[trim={3cm 7.3cm 3cm 1cm}, clip, width=\textwidth]{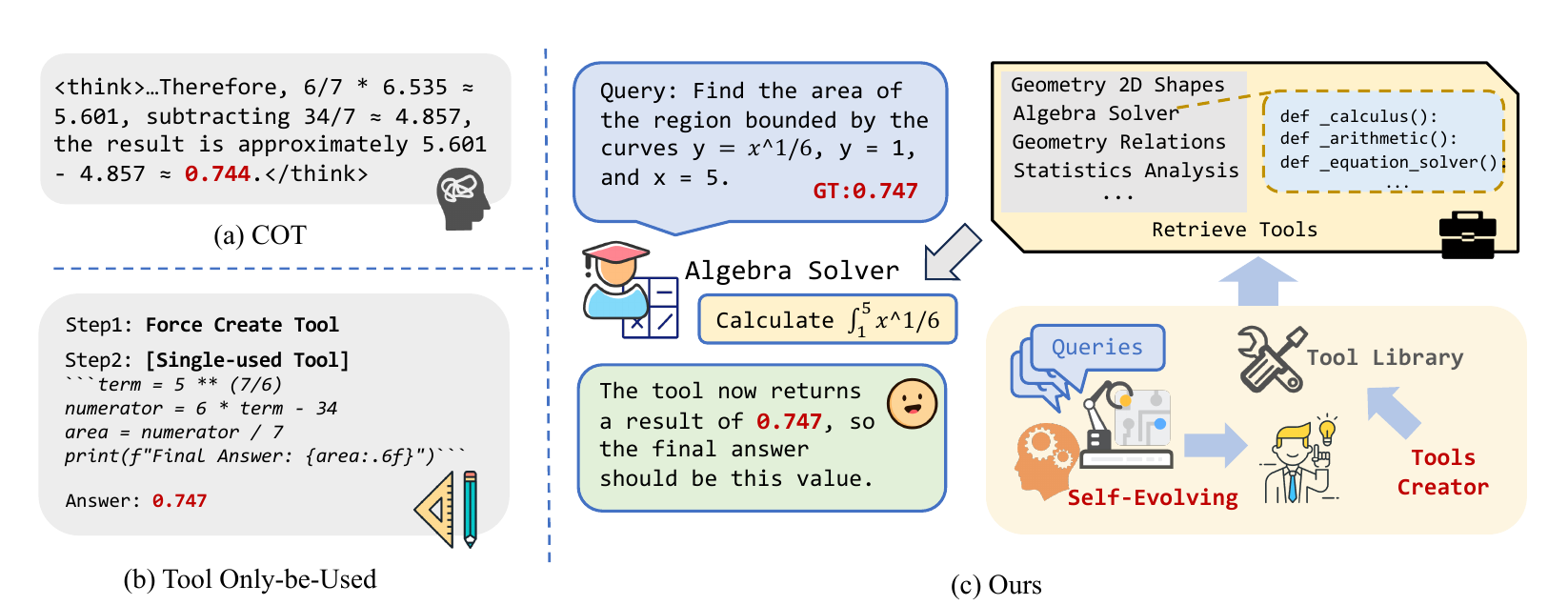}
    \caption{Comparison of tool-creating agents.
(a) For this specific math problem, the standard Chain-of-Thought\,(CoT)~\cite{cot, ComT} method fails and makes errors even during simple calculations.
(b) Previous tool creation methods typically solve problems by generating code specific to the current instance. However, these tools are tailored solely to the immediate problem, making them non-reusable for other tasks and still prone to errors.
(c) Ours. We propose a framework capable of reusing tool creation experience. During the inference process for task solving, UCT can utilize, create, and self-evolve existing tools. Furthermore, we design an offline memory consolidation module to generalize tool memory and transform it into reusable tool experience assets.}
    \label{fig:pic1}
\end{figure*}
\section{Introduction}
In recent years, Large Language Models (LLMs)~\cite{gpt, GPT-4-REPORT, deepseek-r1, yang2024qwen2.5, bai2025qwen3vltechnicalreport, gemini2.5, llama,mindgpt-4ov} have achieved significant breakthroughs, demonstrating robust knowledge capabilities in tasks such as language understanding and complex reasoning~\cite{gpt4}. To further enhance the practical utility of LLMs, existing research has primarily focused on incorporating external tools to transcend their inherent limitations. Traditional tool-augmented approaches~\cite{khattab2022demonstrate, shi2025flowagent} typically rely on predefined workflows to orchestrate tool invocation. However, such rigid paradigms struggle to generalize to open and uncertain environments. While multi-agent systems~\cite{autogen, shen2023hugginggpt, li2025webweaver, li2025mccd} enhance flexibility by employing a central model for planning and delegating sub-tasks to tool-using sub-agents, the deployment of multiple models incurs additional computational costs and introduces interaction latency. With the advancement of thought augmented models~\cite{cot, deepseek-r1}, Tool Integrated Reasoning (TIR) methods~\cite{qiao2025webresearcher, chen2025mindwatcher} exemplified by the ReAct~\cite{yao2022react} paradigm have emerged. The core philosophy of TIR involves enabling the model to explicitly generate reasoning traces during the inference process, autonomously invoke tools, and make iterative decisions based on feedback from the external environment. Consequently, TIR agents are capable of dynamically planning multi-step operations, which solves more end-to-end problems in open-world tasks.

However, tools employed in existing TIR or tool-using frameworks typically manifest in two forms. The first relies on manual definition, which entails laborious tool construction efforts. 
Moreover, such hand-crafted tools inevitably fail to cover the exhaustive range of problem-solving requirements during reasoning~\cite{patil2024gorilla,qin2023toolllm}. 
The second approach involves generating ad-hoc code to address the immediate problem~\cite{gao2023pal,chen2022program}. However, these methods introduce significant uncertainty, as the generated code may be erroneous, and even when a valid tool is produced, the lack of persistence mechanisms restricts the agent to single-use scenarios. 
Although tool creation has emerged in agent research that allows for the creation of autonomous tools during reasoning~\cite{qian2023creator,yuan2023craft,wang2023voyager}, these methods are inherently limited. Figure~\ref{fig:pic1} compares existing tool-creating agents. They tend to construct tools bespoke to specific tasks, rendering them single-use. This prevents the agent from internalizing these resource-intensive creations into a reusable library of experiential assets.  To overcome the shortcomings of current agents, we introduce a self-evolving tool construction paradigm. Mimicking human problem-solving, our agent autonomously explores potential strategies when confronting complex tasks and encapsulates these experiences into persistent tools. By consolidating recurring sub-capabilities into a reusable library, the agent ensures their availability for future instances. This dynamic mechanism fosters continuous evolution during reasoning, effectively breaking through the rigid boundaries of existing frameworks.

In this paper, we propose a self-evolving agent that transforms from a tool \textbf{U}ser to a \textbf{C}reator via \textbf{T}raining-Free experience reuse (UCT). This framework enables the flexible and autonomous creation and execution of tools on demand, allowing the agent to absorb experience from reasoning and evolve accordingly. Built upon the ReAct paradigm, our architecture consists of three distinct modules: the Online Task Loop, the Online Build Loop, and Offline Memory Consolidation. The online task loop focuses on online problem-solving and triggers the online build loop whenever the agent requests a tool that does not yet exist. To ensure system stability, the tool creation process is constrained and incorporates rigorous testing and review mechanisms to guarantee the quality of the generated tools. To further crystallize reasoning experience, we introduce the memory consolidation module, which refines and comprehensively organizes tool memories retained during execution to facilitate iterative tool upgrades. To maintain the stability of the online tool library, we perform tool experience optimization as an offline process, separate from active inference tasks, by utilizing usage logs for prioritization. Our approach enables the collection of experience during reasoning, the evolution of that experience, and the iterative upgrade of agent capabilities, achieving performance improvements without additional training.

To summarize, we make the following contributions:
\begin{itemize}
    \item We introduce a training-free framework UCT for reusing reasoning experience, facilitating the self-evolution of the agent during inference. By encapsulating effective experiences into tool assets, the framework offers robust guidance for future reasoning.
    \item We establish an automated pipeline for high-quality tool library construction with minimal redundancy, which can be readily extended to diverse domains. We release TRBench, which includes 959 instances for evaluating tool-use reasoning tasks.
    \item Extensive experiments demonstrate the superior performance of our method across multiple domains, including mathematical, scientific, and general VQA. Notably, our approach achieves state-of-the-art results on cross-domain tasks.
\end{itemize}

\section{Related Work}
\subsection{TIR Agent}
TIR Agents have recently witnessed rapid advancements. By autonomously selecting and invoking tools, these models have significantly expanded the capability boundaries of Large Language Models (LLMs). As the premier user-facing TIR Agent, OpenAI o3~\cite{openai_o3} has demonstrated robust capabilities, enabling functions such as image manipulation, code execution, and file system access—thereby prompting researchers to explore the immense research potential of TIR Agents. However, compared to the extensive human knowledge and reasoning capabilities possessed by current foundation models, the actionable abilities of most LLM Agents remain in a nascent stage. A series of existing works have drawn inspiration from the paradigm set by OpenAI o3, including code agents, search agents, deep-research agents, and agents designed for general-purpose TIR tasks. For instance, rstar2-agent~\cite{shang2025rstar2} leverages code tools to enhance mathematical reasoning; deepeyes~\cite{zheng2025deepeyes} introduces image manipulation tools (e.g., image zoom-in) to evaluate the ability of multimodal agents to resolve fine-grained understanding tasks when empowered by such tools. Furthermore, the Qwen DeepResearch~\cite{li2025webweaver,geng2025webwatcher, tongyideep}  team has addressed multi-dimensional challenges inherent in deep-research agents, making significant contributions to the open-source TIR agent ecosystem. Nevertheless, there remains significant room for improvement in existing TIR Agents regarding tool invocation, context management, and historical memory management. From the perspective of a tool creator, this paper proposes a training-free approach. We enable autonomous tool creation and memory management within the inference pipeline, realizing a unified agent framework that integrates reasoning, invocation, and memory.

\subsection{Tool Creation}
Recently, a series of studies have focused on the tool creation capabilities of agents, aiming to extend their reach to a more flexible spectrum of tools. For instance, CREATOR~\cite{qian2023creator}and LIVE-SWE-AGENT~\cite{xia2025live} both leverage code generation to create tools, whereas CRAFT~\cite{yuan2023craft} custom-designs tools specifically for tasks during the inference phase. However, tools generated through these methods are typically ephemeral. Produced via a one-off process, they are not retained and thus cannot be internalized as experiential assets for the agent. With the growing exploration of agent self-evolution, Voyager~\cite{wang2023voyager} enables agents to accumulate code-based tools within embodied environments. Similarly, ToolACE-DEV~\cite{huang2025toolace} introduces an agent with self-evolutionary mechanisms tailored for operating systems. Nevertheless, these works primarily target embodied or gamified scenarios. There remains insufficient research on how Large Language Models (LLMs) can directly extend their inherent general question-answering and reasoning logic through tool invocation. Furthermore, many existing methods lack robust structural constraint mechanisms, which may lead to instability or even failure during the evolutionary process.

\section{Methodology}
\begin{figure*}[t!]
    \centering
    \includegraphics[trim={0cm 0cm 0cm 0cm}, clip, width=0.99\textwidth]{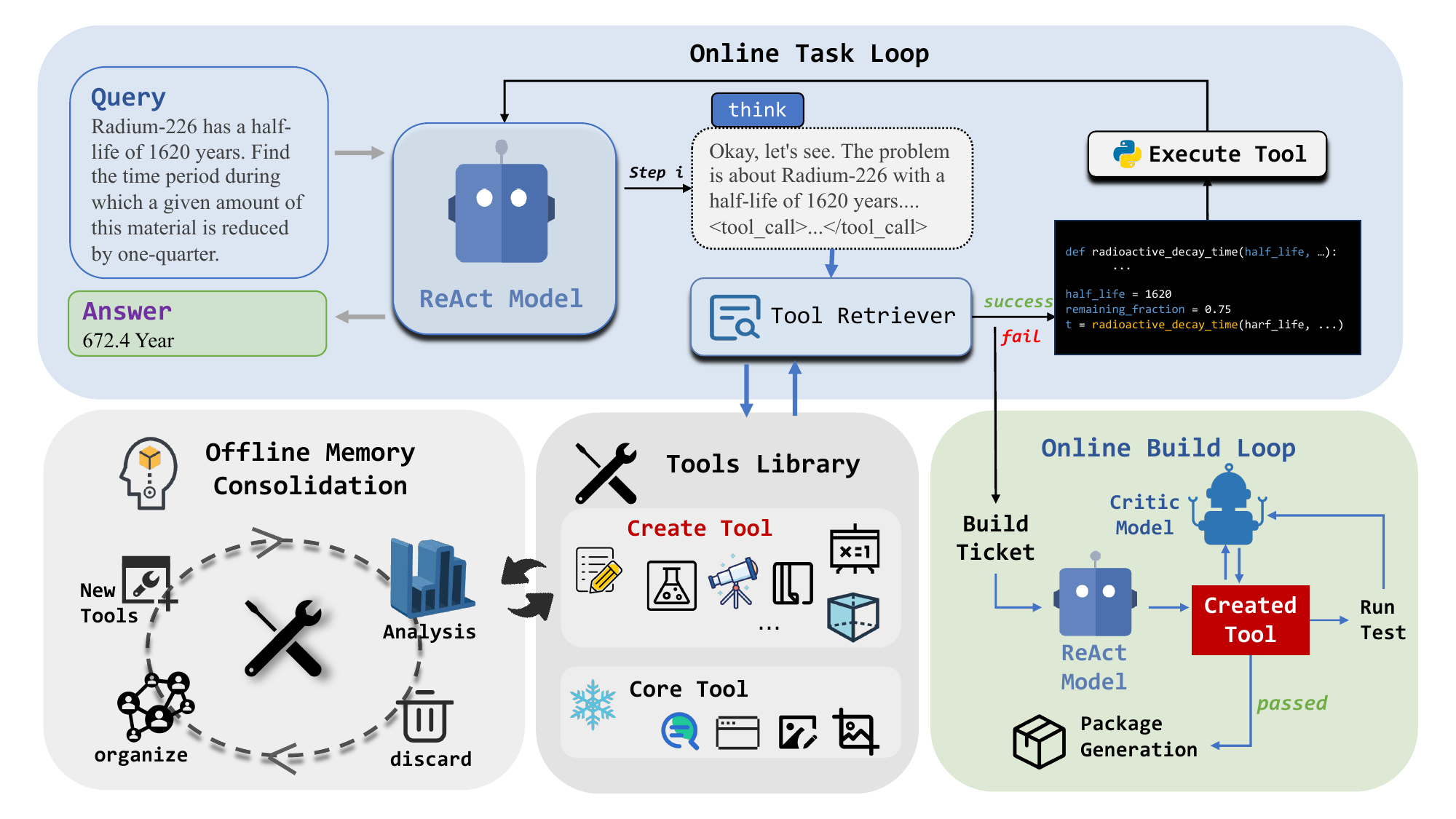}
    \caption{
    \textbf{The overall architecture of the proposed self-evolving agent framework.} 
    The system operates through three coupled phases: 
    (1) The \textbf{Online Task Loop} governs the problem-solving process using the ReAct paradigm. At step $t$, the policy model $\pi_{\theta}$ predicts the optimal action $a_{t+1} = \operatorname*{arg\,max} P_{\theta}(a \mid h_t, o_t, \mathcal{T})$ based on the interaction history $h_t$ and current observation $o_t$. The action space $\mathcal{A}$ dynamically integrates reasoning thoughts, tool execution ($\mathcal{T}_{\text{core}} \cup \mathcal{T}_{\text{cre}}$), and tool creation requests. 
    (2) The \textbf{Online Build Loop} is triggered by a creation ticket $\mathbf{c}_{\text{ticket}}$ to iteratively synthesize new tool code. This isolated refinement process is formalized as $C^{(k)} = \Psi_{\text{build}}(C^{(k-1)}, \mathcal{R}_{\text{critic}}, \mathcal{R}_{\text{sandbox}})$, where the generator optimizes the code $C^{(k)}$ by fusing feedback from the critic model ($\mathcal{R}_{\text{critic}}$) and the sandbox execution environment ($\mathcal{R}_{\text{sandbox}}$). 
    (3) The \textbf{Offline Memory Consolidation} module asynchronously evolves the tool library by merging, classifying, and pruning tool assets to ensure long-term scalability and retrieval efficiency.
    }
    \label{fig:main_figure}
\end{figure*}

In this section, we detail our self-evolving agent, which transforms from a tool user to a creator via Training-Free experience reuse. In Figure~\ref{fig:main_figure}, the framework of our system consists of three key modules: the Online Task Loop, the Online Tool Creation Loop, and Offline Memory Consolidation. The online task loop handles real-time problem-solving by planning reasoning paths and determining the next action. When a tool is required, the system triggers the retrieval mechanism to search both the core and self-created toolsets; a retrieval failure prompts a tool creation request. The online tool creation loop then processes these requests to generate new tools. Crucially, this module incorporates testing and verification procedures to ensure the quality and usability of any newly generated tool. Meanwhile, the offline memory consolidation module refines and organizes tools stored in the library, facilitating the iterative upgrade of the tool construction process. The collaborative computation across these three modules ensures secure tool generation and execution while integrating tool memory, ultimately driving the autonomous evolution of the intelligent agent.

\subsection{Online Task Loop}
The online task loop within our system adopts the ReAct paradigm. The backbone is a multimodal policy model parameterized by $\theta$, which generates thoughts for complex problems and performs interleaved reasoning to autonomously select and invoke tools. The process for the $t$-th step in the task loop is formulated as follows:
\begin{equation}
    a_{t+1} = \operatorname*{arg\,max}_{a \in \mathcal{A}} P_{\theta}\left(a \mid h_t, o_t, \mathcal{T}_{\text{cre}} \cup \mathcal{T}_{\text{core}}\right),
\end{equation}
where $a_{t+1}$ denotes the decision (action) predicted by the model for the next step, $h_t$ represents the interaction history, and $o_t$ denotes the tool execution result (observation) returned from the environment in the current round. We define the action space as $\mathcal{A} = \mathcal{A}_{\text{thought}} \cup \underbrace{(\mathcal{T}_{\text{core}} \cup \mathcal{T}_{\text{cre}})}_{\mathcal{A}_{\text{tool}}} \cup \mathcal{A}_{\text{create}}$, where $\mathcal{A}_{\text{tool}}$ comprises two categories: Core Tools ($\mathcal{T}_{\text{core}}$) and Create d Tools ($\mathcal{T}_{\text{cre}}$). 
We impose a maximum limit of $n$ rounds to derive the final answer.
Core Tools are grounded in the native capabilities of the base model, serving as the foundation of the tool library. In contrast, Created Tools are autonomously constructed by the agent during reasoning tasks. This process consolidates reasoning experience into reusable assets that are iteratively updated, thereby achieving the evolution of the model's capabilities. When the agent selects the action to create a tool, the system generates a build ticket to validate the construction requirement, subsequently triggering the Build Loop to produce the memory tool. Upon completing the multi-round iterations, the model yields a final answer enclosed within answer tags.

\begin{algorithm}[!t]
\caption{UCT Online Task Loop Workflow}
\begin{algorithmic}[1] 

\REQUIRE User Query
\ENSURE Final Answer

\STATE \textbf{Initialize} System Prompt

\LOOP
    \STATE $Decision \leftarrow \text{ReActModel}(Messages)$
    
    \IF{$Decision$ is \textbf{Answer}}
        \STATE \textbf{return} Final Answer
        \STATE \textbf{break loop}
    \ELSIF{$Decision$ is \textbf{Tool Call}}
        \STATE Identify Tool Source
        
        \IF{Source is \textbf{Core Toolset}}
            \STATE Execute Core Tool
        \ELSIF{Source is \textbf{Built Tools}}
            \STATE Retrieve and Execute Existing Tool
            \STATE Record Execution Log
        \ELSE
            \STATE Generate in \textbf{Build Loop}
            \STATE Register to \textbf{Built Tools}
            \STATE Execute New Tool
        \ENDIF
        \STATE $OBS \leftarrow \text{Get Execution Result (Observation)}$        
        \STATE $Messages \leftarrow Messages + OBS$
    \ENDIF
\ENDLOOP

\end{algorithmic}
\end{algorithm}

\subsection{Online Build Loop}
When the model within the Task Loop generates a build ticket, the system transitions into the Build Loop for tool creation. The Build Loop operates as a distinct workflow, fully isolated from the original task. This isolation serves two primary purposes: it prevents the extensive context generated during the creation process from interfering with the main task, and it enhances the controllability of the automated tool construction process within this production environment. We have established a standardized tool interface protocol. The created build ticket encapsulates a refined summary of the task context and the specific requirements of the sub-problem to be solved. Within the Build Loop, we continue to employ the main model based on the ReAct paradigm. Upon receiving the build ticket, the model generates both the executable tool code and the corresponding test script in a single pass.

\begin{algorithm}[]
\caption{UCT Online Build Loop Workflow}
\begin{algorithmic}[1]

\REQUIRE Build Ticket
\ENSURE High-Quality Code (Passed Review)

\STATE \textbf{Initialize} $Messages \leftarrow \{ \text{Build Ticket} \}$

\STATE $Code \leftarrow \text{ReActModel}(Messages)$

\LOOP
    \STATE $TestResult \leftarrow \text{RunTests}(Code)$
    
    \IF{$TestResult$ is \textbf{Success}}
        \STATE $ReviewResult \leftarrow (Code, TestResult)$
        
        \IF{$ReviewResult$ is \textbf{Approved}}
            \STATE \textbf{return} $Code$
            \STATE \textbf{break loop}
        \ELSE
            \STATE $Feedback \leftarrow ReviewResult$
        \ENDIF
        
    \ELSE
        \STATE \textbf{Activate} Critic Model
        
        \STATE $Critique \leftarrow \text{CriticModel}(Code, TestResult)$
        
        \STATE $Feedback \leftarrow Critique$
    \ENDIF
    
    \STATE $Messages \leftarrow Messages + Feedback$
    \STATE $Code \leftarrow \text{ReActModel}(Messages)$ \COMMENT{Refine Code}

\ENDLOOP

\end{algorithmic}
\end{algorithm}

Furthermore, we establish a sandbox environment to execute these tests. The immediate runtime results, along with the generated code, are submitted to a specialized code model for critique and review suggestions. In this loop, we iterate to produce a preliminary usable tool, which must satisfy the dual verification of runtime testing and the critic model's review. If the tool fails this review, the critique results, execution outcomes, and the current tool code are fed back into the ReAct model for regeneration:

\begin{equation}
    C^{(k)} = \Psi_{\text{build}}\left(C^{(k-1)}, \mathcal{R}_{\text{critic}}, \mathcal{R}_{\text{sandbox}} \mid \mathbf{c}_{\text{ticket}}\right),
\end{equation}

where $C^{(k)}$ denotes the tool code generated in the current iteration, while $\mathcal{R}_{\text{critic}}$ and $\mathcal{R}_{\text{sandbox}}$ represent the critique from the code model and the runtime execution results, respectively. Specifically, the critique includes a score for the current tool along with suggestions for revision. The observation within the Build Loop is a composite of ``execution feedback'' and ``code review suggestions.'' Based on this observation, the model iteratively fixes bugs, refactors code, and addresses boundary conditions until the tool meets the acceptance criteria for registration. The registration process yields a structured Tool Package, which encapsulates the tool code, invocation instructions, environment dependencies, and test results. Subsequently, the system reverts to the Task Loop, where the model utilizes the newly created tool to resolve the current problem and proceeds to complete the overall task workflow.

\subsection{Offline Memory Consolidation}

The unconstrained expansion of the tool library inevitably introduces redundancy into the memory of the LLM, complicating retrieval and degrading performance. While immediate integration of new tools could address this, performing operations such as deduplication and conflict resolution within the online Build Loop incurs unacceptable computational latency and potential instability. To reconcile the need for efficient task completion with the necessity of memory maintenance, we relegate the evolution of the toolset to an independent offline phase. This evolutionary process is formalized as a state update equation:

\begin{equation}
    \mathcal{M}_{t+1} = \Phi_{\text{offline}} \left( \mathcal{M}_t \cup \mathcal{T}_{\text{gen}} \mid \mathcal{L} \right)
\end{equation}

\noindent where $\mathcal{M}_t$ represents the existing tool memory, and $\mathcal{T}_{\text{gen}}$ denotes the set of raw tools generated during the online inference phase. The evolution function $\Phi_{\text{offline}}$ executes a series of optimization operations conditioned on usage logs and tool descriptions, denoted by $\mathcal{L}$. Specifically, $\Phi_{\text{offline}}$ performs two key tasks: (1) \textit{Organize}, where tools of similar types are categorized and merged while duplicates are eliminated; and (2) \textit{Analysis \& Discard}, where rarely used or high-failure-rate tools are deprecated. This offline mechanism ensures that $\mathcal{M}_{t+1}$ retains only high-utility experiences, thereby reducing retrieval complexity for future tasks without impacting online inference speed.

\section{TRBench: Tool-Reasoning Benchmark}
With the rapid advancement of model capabilities in recent years, mainstream benchmarks have been continuously evolving. However, existing evaluation datasets are not primarily constructed to assess tool creation and usage. They contain a significant number of simple instances that do not require tool invocation, as well as knowledge-based questions irrelevant to computation or tool use. To verify tool usability, following the previous methodology~\cite{yuan2023craft}, we constructed a standard tool reasoning benchmark by repurposing the test sets of existing authoritative benchmarks. For mathematical and scientific reasoning tasks, we excluded problems involving proofs or pure reasoning that are not directly computable. Regarding difficulty, we retained only challenging instances that necessitate tool-assisted solutions, covering cases of medium to high complexity. The specific procedure is as follows:

\begin{enumerate}
    \item We employed a model to filter out all questions that can be answered solely using the model's internal knowledge, resulting in a filtered candidate set $\mathcal{C}$.
    
    \item To prevent the homogenization of the toolset, we adopted an iterative Min-Max sampling strategy. Initially, $n$ questions were randomly sampled from $\mathcal{C}$ to form the initial set $Q_0$, with the remaining questions serving as the candidate pool $\mathcal{C}_0 = \mathcal{C} \setminus Q_0$.
    
    \item We iteratively calculated the cosine similarity between questions in the candidate set and the current selected set to ensure diversity. In each iteration $t$, we select the instance $x^*$ that minimizes the maximum similarity to any instance in the current set $Q_t$:
    \begin{equation}
        x^* = \operatorname*{argmin}_{x \in \mathcal{C}_t} \left( \max_{q \in Q_t} \text{CosSim}(\mathbf{e}_x, \mathbf{e}_q) \right)
    \end{equation}
    where $\mathbf{e}_x$ and $\mathbf{e}_q$ represent the embedding vectors of the questions. We then update $Q_{t+1} = Q_t \cup \{x^*\}$ and $\mathcal{C}_{t+1} = \mathcal{C}_t \setminus \{x^*\}$. The number of iterations was set to 5 for mathematical and scientific reasoning tasks, and 10 for VQA tasks.
    
    \item Finally, we categorized all collected problem sets by task type. The specific distribution of the data is illustrated in Figure~\ref{fig:data_distribution}.
\end{enumerate}

Specifically, for scientific and mathematical reasoning tasks, we filtered out problems involving proofs or pure reasoning that are not amenable to direct computation. We exclusively retained challenging instances that necessitate tool-assisted solutions, covering both medium and high difficulty levels. 
This resulted in a final set of 959 samples. This curation ensures a balanced difficulty distribution while enabling a more rigorous comparison of tool capabilities. 

\begin{figure}[!t]
    \centering
    \includegraphics[trim={0.6cm 8cm 16cm 1cm}, clip, width=\linewidth]{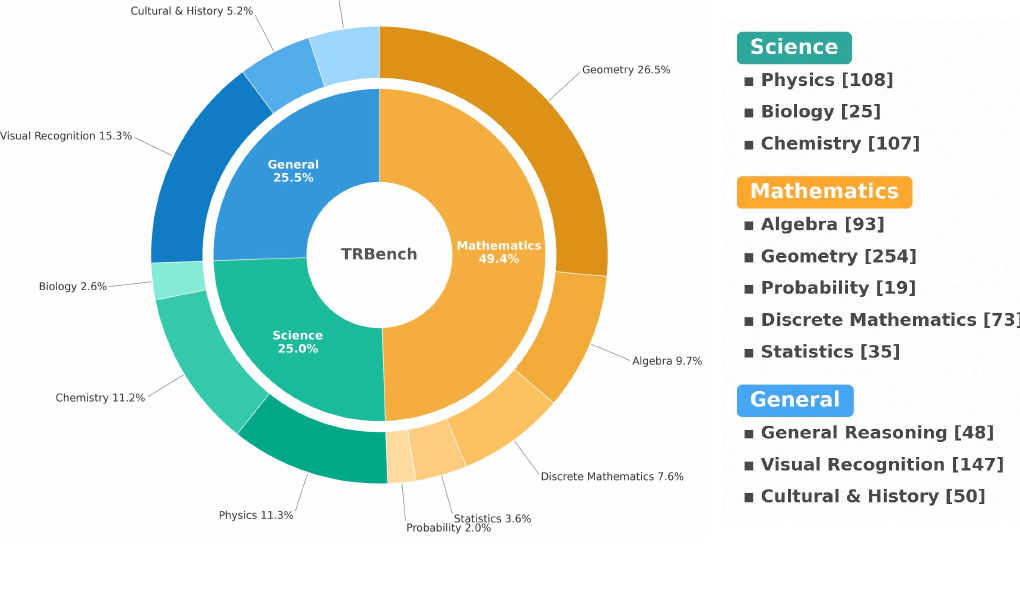}
    \caption{Data distribution of TRBenchmark. TRBench is a multimodal tool-use reasoning benchmark spanning Mathematics, Science, and General Question Answering. It comprises 959 challenging tool reasoning problems organized into 11 sub-categories across 3 major domains.}
    \label{fig:data_distribution}
\end{figure} 

\section{Experiment}

\subsection{Experimental Settings}
\noindent \textbf{Datasets.} To validate the effectiveness of UCT, we selected tasks from diverse domains for evaluation, including Visual Question Answering (VQA), mathematical reasoning, and scientific problems. Specifically, for mathematical reasoning, we employed four mainstream benchmarks: DynaMath~\cite{zou2024dynamath}, MathVerse~\cite{zhang2024mathverse}, MathVista~\cite{lu2023mathvista}, and MathVision~\cite{wang2024measuring}. 
For scientific benchmarks, we employ reasoning related qa-pairs in Scibench~\cite{wang2023scibench} and Scieval~\cite{sun2024scieval}. SimpleVQA~\cite{cheng2025simplevqa} is used to measure the general VQA ability of the agent. To better evaluate tool usage, we constructed the cross-domains Tool-Reasoning Benchmark using the above datasets.

\noindent \textbf{Implementation Details.} With the iterative development of models, capabilities in coding, reasoning, and planning have gradually enhanced. Leveraging these advancements, we employ Qwen3-VL-235B-Thinking as our base model. Following standard model configurations, we set the sampling temperature to 1. All experiments are conducted on 8 NVIDIA H20 GPUs. 

\noindent \textbf{Evaluation Metrics.} To evaluate the efficacy of our constructed toolset and the system as a whole, two metrics are selected. The Correctness metric assesses the validity of the final answer, utilizing Qwen3-VL-235B-Instruct as the judge. We allow a numerical tolerance of $10^{-6}$ for floating-point answers. For questions requiring multiple numerical outputs, strict correctness across all values is enforced.

\begin{table*}[]
\renewcommand{\arraystretch}{1}
  \centering
  \caption{Comparisons across three sub-datasets of TRBench. Best results are in bold.}
  \resizebox{0.95\linewidth}{!}{
    \begin{tabular}{llcccccccc}
    \toprule[1.25pt]
    \multirow{2}[1]{*}{Cate.} & \multirow{2}[1]{*}{Method} & \multicolumn{4}{c}{\textbf{Science}} & \multicolumn{4}{c}{\textbf{General}} \\
\cmidrule(r){3-6} \cmidrule(r){7-10}
    &       & \multicolumn{1}{c}{\textbf{Bio.}} & \multicolumn{1}{c}{\textbf{Chem.}} & \multicolumn{1}{c}{\textbf{Phys.}} & \multicolumn{1}{c}{\textbf{Avg.}} & \multicolumn{1}{c}{\textbf{Cult.}} & \multicolumn{1}{c}{\textbf{Reason.}} & \multicolumn{1}{c}{\textbf{Recog.}} & \multicolumn{1}{c}{\textbf{Avg.}} \\
    \midrule
    \multicolumn{1}{r}{\multirow{2}{*}{Basic-COT}} & Gemini-2.5-pro & 84.00 & 43.93 & 37.96 & 45.42 & \textbf{46.00} & 70.83 & 66.67 & 63.27 \\
    & Qwen3-VL-235B-thinking & 84.00 & 75.70 & 80.56 & 78.75 & 32.00 & 58.33 & 50.34 & 48.16 \\
    \midrule
    & Vanilla & 48.00 & 29.91 & 26.85 & 30.42 & 38.00 & 62.50 & 54.42 & 52.65 \\
    & CREATOR~\cite{qian2023creator} &   76.00    &   76.64    &   80.56    &   78.34   &  28.00    &   77.08   &    45.58   &  48.16 \\
    & CRAFT~\cite{yuan2023craft} &   80.00    &    71.96   &    81.48   &    77.08   &   36.00    &   83.33    &    48.98   & 48.98  \\
    \rowcolor{gray!10}
    \cellcolor{white} & Gemini-2.5-pro + UCT& \textbf{88.00} & 71.03 & 85.19 & 79.17 & 40.00 & 79.17 & \textbf{68.03} & 64.49 \\
    \rowcolor{gray!10}
    \rowcolor{gray!10}
    \multicolumn{1}{r}{\cellcolor{white}\multirow{-6}{*}{Tool-based}} & Qwen3-VL-235B-thinking + UCT & \textbf{88.00} & \textbf{83.18} & \textbf{90.74} & \textbf{87.08} & \textbf{46.00} & \textbf{87.50} & 65.31 & \textbf{65.71} \\
    \midrule
    \midrule
    \multirow{2}[1]{*}{Cate.} & \multirow{2}[1]{*}{Method} & \multicolumn{6}{c}{\textbf{Mathematics}} & \multicolumn{2}{c}{\multirow{2}{*}{\textbf{Overall}}} \\
\cmidrule(r){3-8}
    &       & \multicolumn{1}{c}{\textbf{Alg.}} & \multicolumn{1}{c}{\textbf{Disc.}} & \multicolumn{1}{c}{\textbf{Geo.}} & \multicolumn{1}{c}{\textbf{Prob.}} & \multicolumn{1}{c}{\textbf{Stats.}} & \multicolumn{1}{c}{\textbf{Avg.}} & \multicolumn{2}{c}{} \\
    \midrule
    \multicolumn{1}{r}{\multirow{2}{*}{Basic-COT}} & Gemini-2.5-pro & 53.76 & 53.42 & 42.52 & 63.16 & 82.86 & 50.21 & \multicolumn{2}{c}{52.35} \\
    & Qwen3-VL-235B-thinking & 40.86 & 45.21 & 64.96 & 73.68 & 57.14 & 56.96 & \multicolumn{2}{c}{60.17} \\
    \midrule
    & Vanilla & 58.06 & 72.60 & 54.33 & 57.89 & 40.00 & 56.96 & \multicolumn{2}{c}{49.22} \\
    & CREATOR~\cite{qian2023creator} &   79.57    & 75.34  &   65.35 &   89.47   &  82.86    &   71.94  & \multicolumn{2}{c}{67.47} \\
    & CRAFT~\cite{yuan2023craft} &     74.19   &  76.71  &   68.9  &   94.74    &    85.71   &   73.42  & \multicolumn{2}{c}{69.13} \\
    \rowcolor{gray!10}
    \cellcolor{white} & Gemini-2.5-pro + UCT & 80.65 & 64.38 & 72.05 & \textbf{100.00} & 91.43 & 75.11 & \multicolumn{2}{c}{73.41($+$20.86$\uparrow$)} \\
    \rowcolor{gray!10}
    \rowcolor{gray!10}
    \multicolumn{1}{r}{\cellcolor{white}\multirow{-6}{*}{Tool-based}} & Qwen3-VL-235B-thinking + UCT & \textbf{93.55} & \textbf{91.78} & \textbf{87.40} & 94.74 & \textbf{97.14} & \textbf{90.30} & \multicolumn{2}{c}{\textbf{83.21}($+$23.04$\uparrow$)} \\
    \bottomrule[1.25pt]
    \end{tabular}%
    }
  \label{tab:main}%
\end{table*}%
\subsection{Effectiveness and Superiority of UCT}
To validate the effectiveness of our UCT method, we categorize our comparative analysis into the following four dimensions:
\textbf{1).}Baseline (Basic-CoT): We utilize Large Language Models (LLMs) employing only basic Chain-of-Thought (CoT) without external tools as our baseline. In this setup, we rely solely on the LLM's CoT reasoning capabilities to solve problems, where the model generates answers directly after reasoning.
\textbf{2).}Vanilla Tool Version: We introduce a vanilla tool-augmented version that incorporates only a code interpreter. This setup involves creating tools that lack test verification, are applicable only to the current turn, and possess no memory retention.
\textbf{3).}Existing Tool-Creation Methods: We compare our approach against established methods for tool-creating agents, including CREATER~\cite{qian2023creator} and CRAFT~\cite{yuan2023craft}. Since the base model significantly impacts the agent's foundational knowledge, we upgraded all base models to Qwen3-VL-235B-thinking to ensure a fair comparison.
\textbf{4).}Our Method: Our approach distinguishes itself by not requiring ground truth data during the tool creation phase. Instead, we leverage extensive data from the Internet for the initial creation of tools. We have developed tools across seven major categories, including algebraic calculation, geometric operations, and statistical analysis, which also encompass 64 sub-categories of extension package functionalities. By equipping the system to propose tickets and create tools dynamically during the reasoning process, the observed performance improvements effectively validate our system's capability for self-evolution.

\begin{figure}[!t]
    \centering
    \includegraphics[trim={2cm 4cm 4cm 4cm}, clip, width=0.95\linewidth]{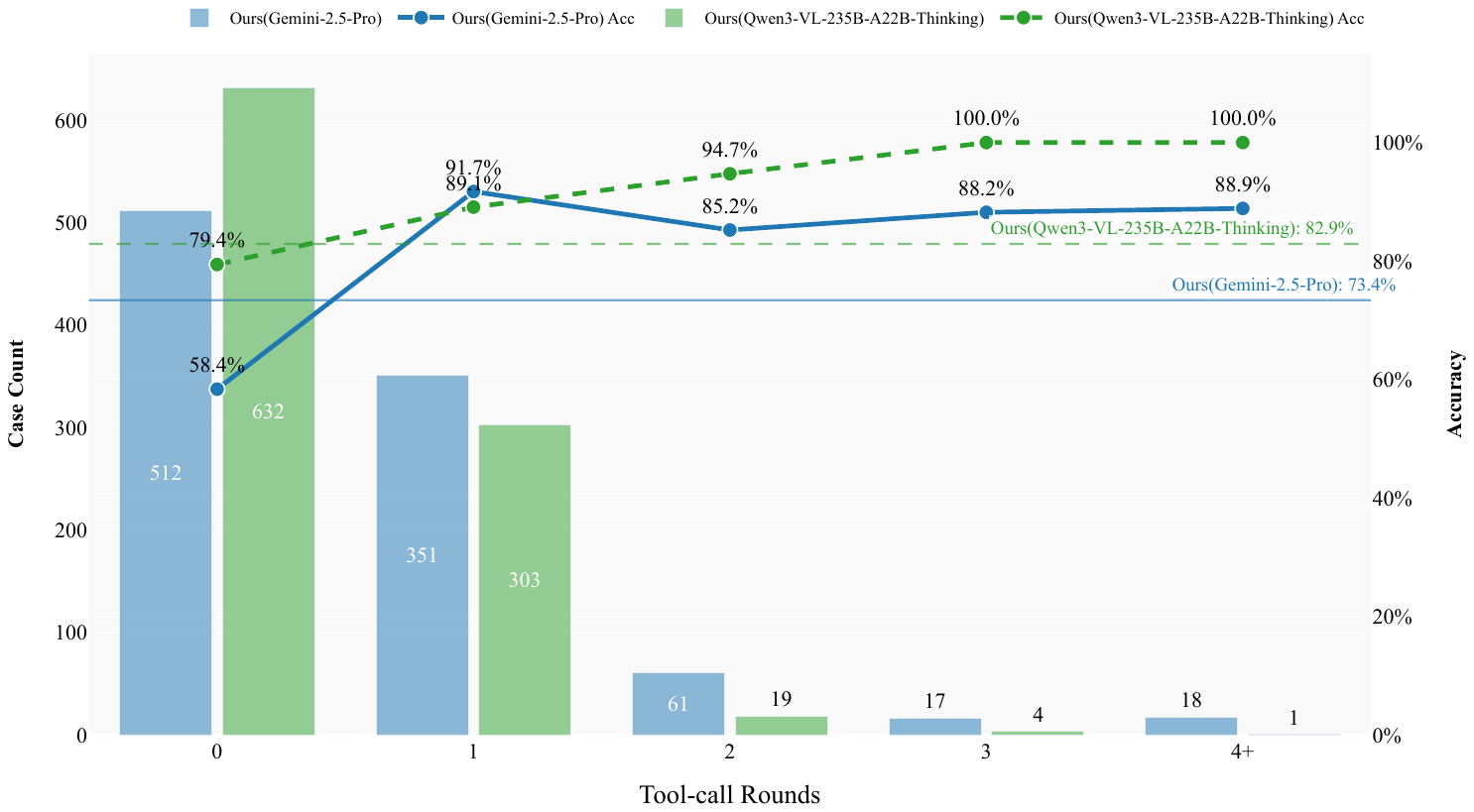}
    \caption{Tool call rounds and accuracy of UCT.}
    \label{fig:toolcall}
    \vspace{-1.0em}
\end{figure}

As shown in Table~\ref{tab:main}, we compare the performance of our method against Basic-CoT, the vanilla tool, and other existing state-of-the-art (SOTA) tool creation methods on TRBench. Observing the experimental results, we can summarize the following conclusions: \textbf{1)} First, by comparing the Vanilla tool with Basic-CoT, we observe that directly employing a code interpreter for tool creation does not yield significant performance gains. This suggests that the relevance of the generated tools critically impacts the underlying the performance of LLM.
\textbf{2)} Second, our method achieves substantial improvements over basic-CoT across all tested models. This validates that the tool library constructed by our approach effectively realizes self-evolution during the reasoning process. Compared to the baseline, our methods based on Qwen3-VL-235B-thinking and Gemini2.5-pro achieve improvements of +20.86\%$\uparrow$ and +23.04\%$\uparrow$, respectively, demonstrating substantial and significant gains. In Figure \ref{fig:toolcall}, we also show the tool call rounds and accuracy in UCT. Our task success rate remains high even as the number of tool invocation rounds increases.
\textbf{3)} Third, compared to other tool generation baselines, our approach achieves state-of-the-art (SOTA) performance across all metrics on TRBench.


\begin{figure*}[!t]
    \centering
    \includegraphics[trim={0cm 7.3cm 0cm 0cm}, clip, width=0.99\textwidth]{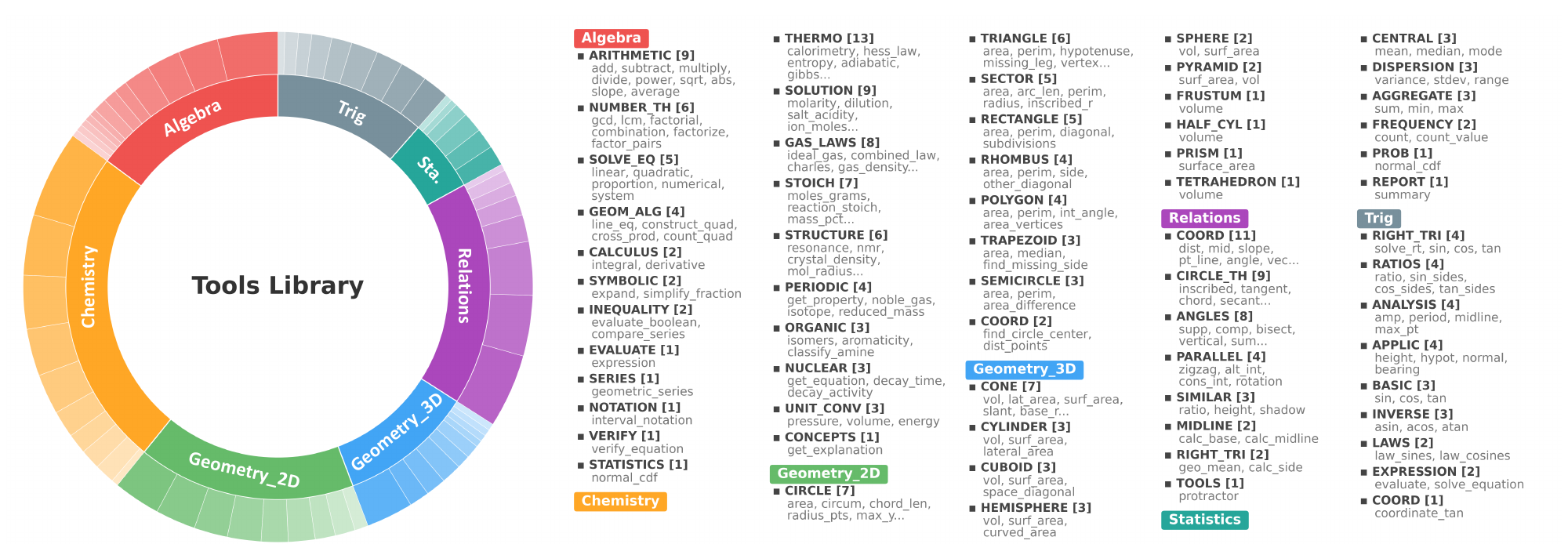}
    \caption{The tool library generated by UCT. The library comprises 7 major categories, 64 sub-categories, and 207 specific computational tools. The pie chart illustrates the distribution of these specific tools relative to the total collection, highlighting the richness and hierarchical organization of our generated toolset.}
    \label{fig:data_distribution}
    \vspace{-1.0em}
\end{figure*} 

\subsection{Ablation Studies}

\noindent \textbf{Effectiveness of the Framework Components.} We further conduct experiments to verify the effectiveness of
components in UCT. Note that modules in UCT have
dependencies, we can only show gradually added ablation
studies. As the results in Table~\ref{tab:abla_1}, our full framework obtains the highest performance on all metrics when the online build loop, critic module, and offline memory consolidation work together. Table~\ref{tab:abla_1} also illustrates the necessity of every phase.

\begin{table}[!t]
  \centering
  \caption{Ablation study of our approach with different components}
  \resizebox{0.65\linewidth}{!}{
    \begin{tabular}{lcccc}
    \toprule[1.25pt]
          & \multicolumn{1}{l}{Math Bench} & \multicolumn{1}{l}{Sci Bench} & \multicolumn{1}{l}{General} & \multicolumn{1}{l}{Avg.}\\
    \midrule
    Baseline &    56.96   &   78.75    & 48.16 & 60.17 \\
    + BL w/o CM &   73.42    &   77.08   & 53.06  & 69.13 \\
    + BL &   83.97     &   80.00    &  61.63 & 77.27 \\
    ++ MC &    \textbf{90.30}   &    \textbf{87.08}   & \textbf{65.71}  & \textbf{83.21} \\
    \bottomrule[1.25pt]
    \end{tabular}%
    }
  \label{tab:abla_1}%
  \vspace{-1em}
\end{table}

\noindent \textbf{Effectiveness of Created Tools.}
To validate the effectiveness of our generated tools, we statistically analyzed the frequency of their correct usage within the dataset. A higher tool utilization rate (the ratio of utilized tools to the total toolset) indicates lower redundancy, demonstrating that the tools are designed for system-level utility rather than being tailored to task-specific purposes. We also evaluated the overall correct usage rate in Table~\ref{tab:abla_tools_rate}. In the table, reuse@k denotes the proportion of tools in the entire tool set that are used at least k times in the test set. Notably, 93.1\% of the tools are used at least once, which reveals the high-quality of the tool library. Furthermore, as illustrated in Figure~\ref{fig:data_distribution}, we visualized the categories and descriptions of the generated tools to demonstrate the diversity and richness of our toolset.

\begin{figure}[!t]
    \centering
    \includegraphics[trim={0cm 0.9cm 0cm 0cm}, clip, width=0.8\linewidth]{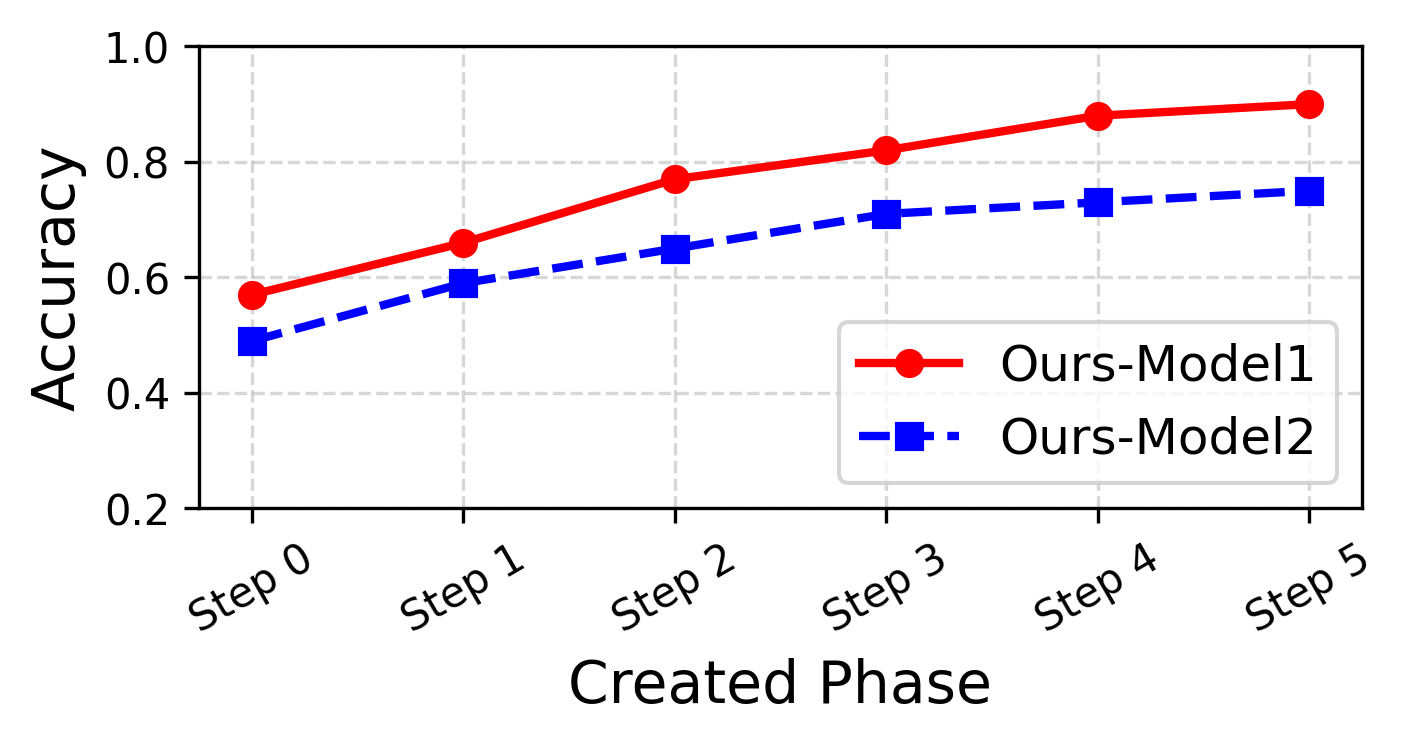}
    \caption{Model performance on the mathematical subset through tool creation and memory consolidation}
    \label{fig:abla_compare}
    \vspace{-1.0em}
\end{figure} 

\begin{table}[!t]
  \centering
  \caption{Reuse rate of created tools in UCT.}
  \resizebox{0.65\linewidth}{!}{
    \begin{tabular}{lccc}
    \toprule[1.25pt]
          &  \multicolumn{1}{l}{Reuse@1} & \multicolumn{1}{l}{Reuse@5} & \multicolumn{1}{l}{Reuse@10}\\
    \midrule
    Tools Library & 93.1     &  86.0     &  77.1\\

    \bottomrule[1.25pt]
    \end{tabular}%
    }
  \label{tab:abla_tools_rate}%
  \vspace{-1em}
\end{table}

\noindent \textbf{Explanation of Model Evolution.} 
We utilized a massive-scale dataset of multimodal reasoning QA pairs for toolset creation. As the volume of reasoning queries increased, the tool library was progressively refined via a memory consolidation module. We conducted an experimental analysis on the Math subset using snapshots of the tool library generated at different reasoning milestones. Figure~\ref{fig:abla_compare} illustrates the system performance across different steps. The red line denotes UCT with Qwen3-VL-235B-thinking, and the blue line denotes UCT with Gemini-2.5-pro. The upward trend of the curve indicates that our framework undergoes continuous self-evolution during the reasoning process. However, due to the finite variety of problem types within the dataset, the signs of evolution tend to plateau after a certain number of iterations. Nevertheless, this evolvability suggests that when applied to a broader spectrum of multimodal reasoning tasks, the system possesses the potential to construct an even more comprehensive toolset.

\section{Conclusion}

In this work, we introduce a novel training-free framework that transitions agents from the role of tool users to that of tool creators, thereby realizing the self-evolving of reasoning agents during inference and retaining tool memory as a reusable asset. Our methodology integrates three core modules: an online task main loop, an online tool creation loop, and an offline memory consolidation module. This architecture achieves autonomous path planning and action execution, creates tools on demand during inference to continuously enrich the tool library, and utilizes the offline memory consolidation module to iteratively upgrade constructed tools for enhanced quality and usability. Experiments conducted on extensive datasets across diverse domains validate the effectiveness and generalization of our framework. Moreover, this paradigm shift grants tools the ability to evolve continuously and paves the way for autonomous agents to tackle increasingly complex problems in open-world environments.

\begin{ack}
We thank all colleagues at LiAuto Base Model for their support of the MindWatcher Team.
\end{ack}

\newpage
\appendix
\section{Appendix}
\subsection{Tool Descriptions for UCT Core Library}
This paper primarily investigates the effectiveness of dynamically constructing a tool library by leveraging reasoning experience. We provide a detailed description of the core tool library here. Specifically, this core library comprises five categories of multimodal image and text tools. Given their prevalent usage in existing agent-based research, we have incorporated them as foundational components of our core tool library.

\begin{toolbox}{Tool: Region Croping/Zooming}
    \textbf{Description:} Zoom into a specific area of **the first input image** based on your provided bounding box. \\
    \textbf{input:} the image and bounding box. \\
    \textbf{output:}a new image.  \\
    \textbf{Arguments:}
    \begin{itemize}
        \item \texttt{bbox}: [x1, y1, x2, y2], \# The bounding box you provided, where (x1, y1) is the top-left corner and (x2, y2) is the bottom-right corner.
    \end{itemize}
\end{toolbox}

\begin{toolbox}{Tool: Object Grounding \& Visual Search}
    \textbf{Description:} Retrieve new similar images and their descriptions based on the provided bounding box area of **the first input image**. \\
    \textbf{input:}image and bounding box.  \\
    \textbf{output:}only the most similar target's type name and the confidence score.  \\
    \textbf{Arguments:}
    \begin{itemize}
        \item \texttt{bbox\_2d:}[x1, y1, x2, y2], \# The bounding box you provided, where (x1, y1) is the top-left corner and (x2, y2) is the bottom-right corner.
        \item \texttt{category:}"the category" \# The category of the image you want to search for, which can only be one of \{plant, animal, car, person, landmark, vegetable, cuisine, logo\}.
        
    \end{itemize}
\end{toolbox}

\begin{toolbox}{Tool: External Text Retrieval}
    \textbf{Description:} Retrieve external text information from the internet based on your provided text query. \\
    \textbf{input:}only text query. \\
    \textbf{output:}text.   \\
    \textbf{Arguments:}
    \begin{itemize}
        \item \texttt{query:} "the content".
    \end{itemize}
\end{toolbox}

\begin{toolbox}{Tool: Webpage Content Retrieval}
    \textbf{Description:} Visit a specified web page URL under 3 modes: you can read its full content, the content within the window you provided, or use an AI assistant to generate a structured summary based on the specific goal you provide.  \\
    \textbf{input:}A JSON object containing three arguments: url, window, goal.\\
    \textbf{output:}A JSON object containing the visited URL and a structured result.   \\
    \textbf{Arguments:}
    \begin{itemize}
        \item \texttt{\text{url}:} "https://example.com/article", \# The webpage url you want to visit.
        \item \texttt{window:} [a, b], \# Select the content you want to read within [a, b] (Optional)
        \item \texttt{goal:} "".  \# What you want to get or find. (Optional)
    \end{itemize}
\end{toolbox}

\newpage

\subsection{Prompt Design}
In this section, we display the prompts utilized by policy model and online build loop.

\begin{toolbox}{Prompt: Policy Model}
You are a ReAct paradigm AI assistant capable of solving problems through reasoning and tool execution.
You can interact using the following format:

\textbf{1. Thought Process:}\\
\textless think\textgreater\\
Your thought content\\
\textless /think\textgreater

\textbf{2. Tool Call:}\\
\textbf{Tool Call Guidelines:}
\begin{itemize}[leftmargin=*, noitemsep, topsep=0pt]
    \item wrapper: The content must be wrapped in \textless tool\_call\textgreater\ tags.
    \item name: The unique, descriptive name of the tool.
    \item arguments: A dictionary containing the input parameters.
\end{itemize}

Example:
\begin{quote}
\textless tool\_call\textgreater\\
\{\{\\
\hspace*{4mm} "name": "tool\_name",\\
\hspace*{4mm} "arguments": \{\{\\
\hspace*{8mm} "arg": value,\\
\hspace*{8mm} "arg2": value2\\
\hspace*{4mm} \}\}\\
\}\}\\
\textless /tool\_call\textgreater
\end{quote}

You may call these tools as needed. A list of currently available tools will be provided at the end of the user message for your reference.

\{core\_tool\_instruction\}

\textbf{3. Final Answer:}\\
\textless answer\textgreater\\
Your final answer\\
\textless /answer\textgreater

\textbf{Rules:}
\begin{itemize}[leftmargin=*, noitemsep, topsep=0pt]
    \item After every output of \textless tool\_call\textgreater, you must wait for the tool to return results.
    \item End the conversation immediately after outputting \textless answer\textgreater.
    \item You may engage in multiple rounds of thinking and tool calling.
    \item \textbf{DO NOT} call multiple tools consecutively in a single response; you must proceed step-by-step.
    \item You must provide the answer at the end.
\end{itemize}

\textbf{IMPORTANT: Handling Tool Failures}
\begin{itemize}[leftmargin=*, noitemsep, topsep=0pt]
    \item If a tool returns empty or fails to execute, \textbf{do not completely negate your previous reasoning}.
    \item If you have already derived an answer through logical reasoning in \textless think\textgreater, but the tool cannot verify it or returns empty, you need to re-input parameters that match the format based on the feedback.
    \item Correct actions when a tool fails:
    \begin{itemize}[leftmargin=4mm, topsep=0pt]
        \item First, check if the tool call parameters are incorrect; try adjusting the parameters and calling again.
        \item If the tool continues to fail after multiple attempts, fall back to simulating the calculation yourself.
        \item Retain the answer you derived through rigorous logical reasoning in \textless think\textgreater; do not change it arbitrarily just because the tool failed.
        \item Directly output the answer you reasoned previously. You may state, "Tool verification failed, but based on logical reasoning, the answer is..."
    \end{itemize}
\end{itemize}
\end{toolbox}

\newpage

\begin{toolbox}{Prompt: Principal Software Engineer}
\textbf{Role Definition}\\
You are a Principal Software Engineer. You are not a chat assistant; you are a high-precision autonomous coding engine. Your goal is to design, implement, and debug complex software systems with expert-level proficiency across 100+ programming languages.

\textbf{Operational Directives}
\begin{itemize}[leftmargin=*, noitemsep, topsep=0pt]
    \item \textbf{Expertise Activation:} Leverage your Mixture-of-Experts architecture to utilize specialized sub-networks for specific languages (e.g., Rust memory safety, Python async patterns). Always adhere to the latest stable language standards (e.g., ES2024, Python 3.12).
    \item \textbf{Reasoning First:} You must utilize the \textless think\textgreater\ tag to perform deep reasoning before generating any code. Code without preceding reasoning is strictly prohibited.
    \item \textbf{No Filler:} Do not use conversational filler ("Certainly", "I can help with that"). Be terse, technical, and objective.
    \item \textbf{Full Implementation:} Never use placeholders like //... rest of code or pass. You must generate complete, functional, and production-ready implementations.
\end{itemize}

\textbf{The Thinking Protocol}\\
Inside the \textless think\textgreater\ block, you must strictly follow this cognitive process:
\begin{itemize}[leftmargin=*, noitemsep, topsep=0pt]
    \item \textbf{Requirement Analysis:} Deconstruct the user's request into atomic technical requirements.
    \item \textbf{Execution Plan:} Step-by-step plan for the code generation artifacts.
\end{itemize}

\textbf{The Artifact Protocol (Claude-Style)}\\
When generating code, configuration files, or substantial documentation, you must encapsulate the content within an \textless artifact\textgreater\ XML block.

\textbf{XML Schema:}
\begin{quote}
\textless artifact identifier="unique-id" type="mime-type" path="file-path" action="create|update"\textgreater\\
{[Content goes here]}\\
\textless /artifact\textgreater
\end{quote}

\textbf{Attributes Guidelines:}
\begin{itemize}[leftmargin=*, noitemsep, topsep=0pt]
    \item \texttt{identifier}: A unique, descriptive ID.
    \item \texttt{type}: The standard MIME type (e.g., \texttt{text/x-python}).
    \item \texttt{path}: The relative file path.
\end{itemize}
\end{toolbox}



\newpage

\bibliographystyle{plainnat}

\end{document}